\documentclass[10pt,twocolumn,letterpaper]{article}

\usepackage{iccv}
\usepackage{times}
\usepackage{epsfig}
\usepackage{graphicx}
\usepackage{amsmath}
\usepackage{amssymb}
\usepackage{commath}
\usepackage{xcolor}
\usepackage{booktabs}
\usepackage{paralist}
\usepackage{bbm}
\usepackage{multirow}
\usepackage{xspace}
\usepackage{enumitem}
\usepackage{bbm} 
\usepackage{siunitx}
\usepackage{textcomp}
\usepackage{adjustbox}
\usepackage{cellspace} 
\usepackage{algorithm}
\usepackage{algorithmic}
\usepackage{multirow, makecell}

\makeatletter
\DeclareRobustCommand\onedot{\futurelet\@let@token\@onedot}
\def\@onedot{\ifx\@let@token.\else.\null\fi\xspace}
 
\def\ie{\emph{i.e}\onedot} 
\def\cf{\emph{c.f}\onedot} 
  
\def\wrt{\emph{w.r.t}\onedot} 
\def\etal{\emph{et al}\onedot}
\makeatother
\def\fig#1{Fig.~\ref{fig:#1}}
\def\tab#1{Table~\ref{tab:#1}}

\usepackage[pagebackref=true,breaklinks=true,letterpaper=true,colorlinks,bookmarks=false]{hyperref}

\DeclareMathOperator*{\argmin}{\arg\!\min}

\iccvfinalcopy 

\def\iccvPaperID{*} 

\ificcvfinal\fi

\begin{document}

\title{Continual Learning for Image-Based Camera Localization}

\author{Shuzhe Wang\thanks{The first two authors contributed equally.} \quad
Zakaria Laskar\footnotemark[1] \quad
Iaroslav Melekhov \quad
Xiaotian Li \quad
Juho Kannala \\
Aalto University \quad \\
\texttt{firstname.lastname@aalto.fi} \quad \\
}

\maketitle
\thispagestyle{empty}

\begin{abstract}

   For several emerging technologies such as augmented reality, autonomous driving and robotics, visual localization is a critical component. Directly regressing camera pose/3D scene coordinates from the input image using deep neural networks has shown great potential. However, such methods assume a stationary data distribution with all scenes simultaneously available during training. In this paper, we approach the problem of visual localization in a continual learning setup -- whereby the model is trained on scenes in an incremental manner. Our results show that similar to the classification domain, non-stationary data induces catastrophic forgetting in deep networks for visual localization. To address this issue, a strong baseline based on storing and replaying images from a fixed buffer is proposed. Furthermore, we propose a new sampling method based on coverage score (Buff-CS) that adapts the existing sampling strategies in the buffering process to the problem of visual localization. Results demonstrate consistent improvements over standard buffering methods on two challenging datasets -- 7Scenes, 12Scenes, and also 19Scenes by combining the former scenes\footnote {Code and materials are available at \url{https://github.com/AaltoVision/CL_HSCNet}.}.
\end{abstract}

\section{Introduction}
\label{sec:intro}

Camera relocalization is a fundamental problem aimed at estimating 6 degree-of-freedom (DoF) camera pose with respect to a known environment. Visual localization aims to solve this problem requiring only RGB images as input~\cite{SattlerLK11,sattler2012improving,SattlerLK17,Sattler17cvpr}. Traditional methods~\cite{sarlin2019coarse,sarlin2018leveraging,SattlerLK11,sattler2012improving,SattlerLK17,Sattler17cvpr,inloc,Svarm2017,sattler2018benchmarking} require building a 3D map of the environment followed by an explicit matching stage~\cite{schoenberger2016sfm,schoenberger2016mvs} to establish 2D pixels to 3D coordinates. Recently with the success of deep neural networks, the problem can now be solved end-to-end by directly regressing the camera pose~\cite{mapnet2018,KendallC15bay,Kendall_2017_CVPR,MelekhovYKR17,Walch_2017_ICCV,Sattler2019,radwan2018vlocnet++,valada2018deep, kendall2015convolutional} or 3D scene coordinates~\cite{Brachmann_2019_ICCV,Brachmann_2019_ICCV_NG,brachmann2020visual,li2020hierarchical,Yang_2019_ICCV}. This has shown to be more accurate than feature-based methods (at least for small scale environments).

\begin{figure*}
\begin{center}

   \includegraphics[width=0.85\linewidth]{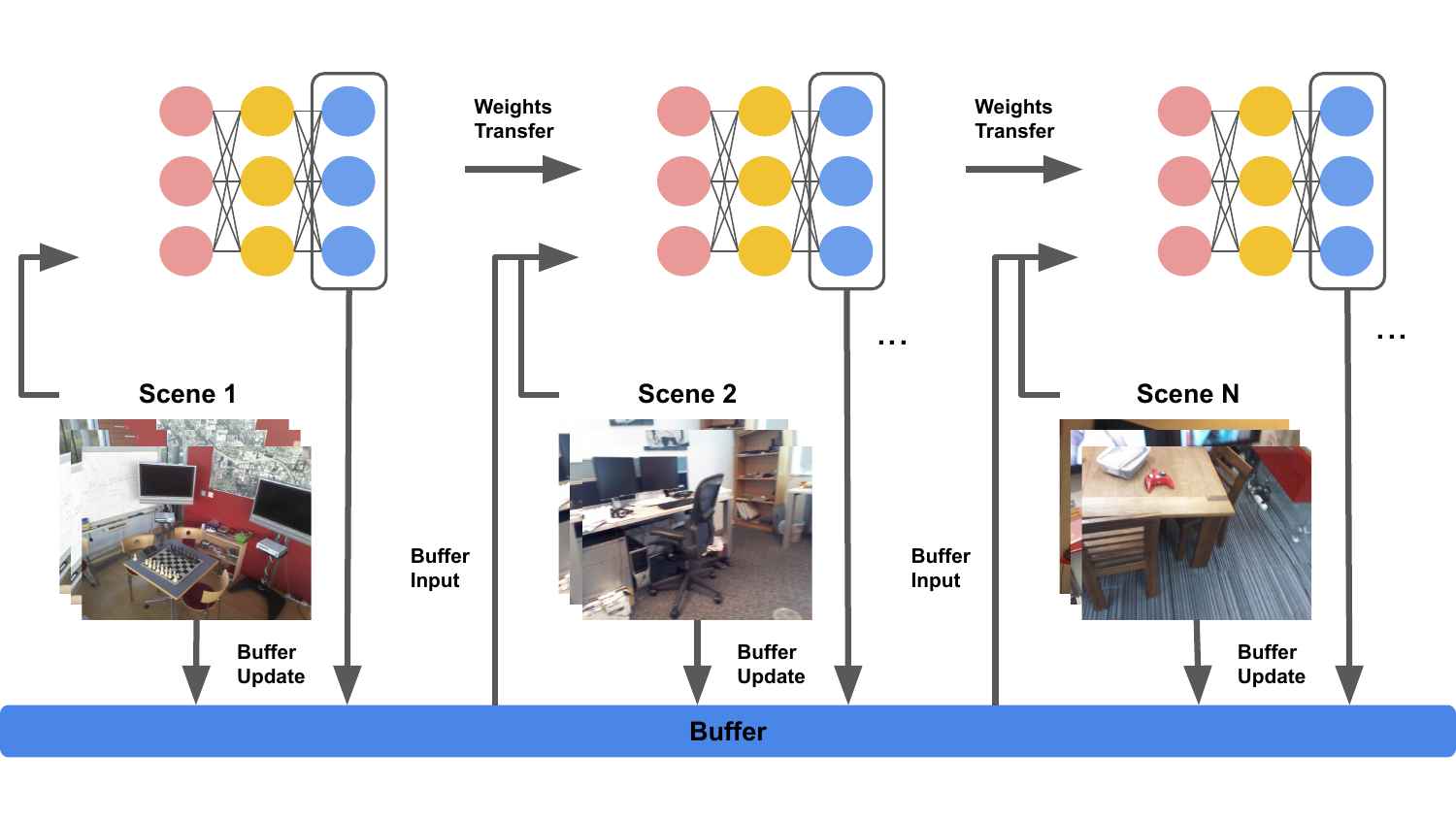}
   \end{center}
   \vspace{-10px}
   \caption{Overview of our replay-based continual learning approach for visual localization. During each scene (task) iteration, the model is updated using the current and previous task samples. The former is sampled from a small fixed-size buffer. After the training is over, a small subset of the current task samples are stored in the buffer by replacing parts of stored data from previous tasks.
\label{fig:main structure}}
\end{figure*}

One of the limitations of end-to-end regression methods for visual localization is limited scalability to larger environments with several scenes. Although the methods performed well when trained and evaluated on a single scene, the performance quickly degraded when jointly trained on multiple scenes. This was mitigated by considering a hierarchical approach~\cite{li2020hierarchical} to localize a given input image -- first obtain a coarse localization in terms of the scene or sub-scene, followed by estimating a finer camera pose estimate. In this work, we push the methods further towards a general intelligence setting -- learn continually from the incoming stream of data. Under this setting, all the scenes are not available during training but encountered sequentially one after the other as shown in Figure~\ref{fig:main structure}. There are several benefits in terms of sample and memory efficiency to learning tasks in a continual manner over the setting of training jointly over all tasks. In the joint training setting, each time a scene has changed, the model needs to be retrained on all the scenes in the database -- even the ones that have not undergone any change. Adding new scenes to the database also requires model retraining which affects the scalability. Due to the above issues, the full dataset needs to be stored in memory.
In contrast, continual learning (CL)~\cite{aljundi_2017,aljundi_2018,chaudhry2021using,kirkpatrick_2016} aims to reduce the computational costs by fine-tuning the model only on the changed/new scene and images from the previous scenes stored in a small buffer. Furthermore, the memory costs are also reduced as only the data for the current scene needs to be stored in memory along with a small buffer of images from previous scenes. This is of particular importance for mobile applications where storage capacities are device constrained.

Solely training on the images from the current scene leads to catastrophic forgetting of knowledge gained from prior scenes. This is attributed to the interference of the gradients from the current task images with the model parameters learned on previous scenes. The performance of neural networks in such non-stationary data distribution setting is well studied under the domain of continual learning. The CL problem is broadly categorized into i) \textit{class/task} CL: all the data from current class/task is available and the model is allowed to have repeated passes through the whole dataset, and ii) \textit{online} CL: the task boundary changes suddenly. To mitigate the challenges of CL, several approaches are proposed: i) \textit{regularization} methods~\cite{kirkpatrick_2016, aljundi_2018, nguyen_2018} that penalize changes in weights considered important for previous scenes or directly impose orthogonality constraints in the training objective, ii) \textit{modular} methods~\cite{aljundi_2017, rosenbaum_2018, chang_2018} that increase the model capacity assigning new parameters for each task, and iii) \textit{replay} methods~\cite{rebuffi_2017,chrysakis2020online} that perform experience replay by storing samples from previous scenes in a fixed size buffer or using generative models to generate images of past scenes. All the three methods incur limited memory and computational costs as regularization methods require past gradients or feature maps to be stored in memory like the replay-based approaches, while modular approaches require an increase in model size. For a fixed model size, experience replay based methods have shown superior performance compared to regularization methods, and in some recent works~\cite{chaudhry2021using}, a combination of both also leads to good results.

In this work, we consider the \textit{task} CL setting for the problem of visual localization in the context of experience-replay based solutions. We adopt some of the buffering methods from literature -- \textit{Reservoir}~\cite{vitter1985random}, and \textit{Class-balance}~\cite{chrysakis2020online} to perform experience replay. A strong baseline is created using these methods and challenges specific to the visual localization problem highlighted. Unlike the classification domain where each image is representative of the whole class, visual localization scenes consist of diverse sets of images spanning a whole 3D environment. Existing buffering methods do not take the 3D scene layout into account. Storing images from just one part of the scene does not guarantee generalization to images from other disjoint parts. To retain performance on several parts of the scene, we propose a buffering process that ensures the images stored in the buffer will have higher scene coverage. This is done by computing a coverage score factor that indicates if buffering new incoming image will improve the existing coverage score of buffer images. The proposed buffering algorithm outperforms existing methods on challenging datasets -- 7Scenes, 12Scenes, and also 19Scenes obtained by combining the former scenes.

To summarize, we make the following contributions:

\begin{itemize}

    \item
    Introduce the problem of continual learning for visual localization.
    \item 
    Create a strong experience-replay baseline from existing buffering methods across several indoor datasets.
    \item
    Propose a new buffering strategy conditioned on the 3D geometry of the scene.
    
\end{itemize}

\section{Related Work}
\label{sec:related}

\noindent \textbf{Visual localization.}
Visual localization is the task of estimating 6-DoF camera pose from an image. Conventional methods~\cite{sarlin2019coarse,sarlin2018leveraging,SattlerLK11,sattler2012improving,SattlerLK17,Sattler17cvpr,inloc,Svarm2017,sattler2018benchmarking} solve it by matching image features against a prebuilt 3D map~\cite{schoenberger2016sfm,schoenberger2016mvs}. The camera pose can then be recovered from the 2D-3D matches in a RANSAC~\cite{RANSAC} optimization loop. Recently with the success of deep neural networks,  learning-based methods have been proposed to tackle the problem. To learn the entire localization pipeline end-to-end, PoseNet~\cite{kendall2015convolutional} was first proposed to directly regress the absolute camera pose from an RGB image, and was later improved upon and studied in~\cite{mapnet2018,KendallC15bay,Kendall_2017_CVPR,MelekhovYKR17,Walch_2017_ICCV,Sattler2019,radwan2018vlocnet++,valada2018deep}. Instead of  directly regressing the absolute pose, in ~\cite{Balntas_2018_ECCV,Ding_2019_ICCV,LaskarMKK17,zhou2020learn}, neural networks are trained to predict query pose relative to the database images of which the poses are known. Scene coordinate regression methods~\cite{BrachmannMKYGR16,cavallari2019real,CavallariGLVST17,Guzman-RiveraKGSSFI14,Massiceti2017,meng2017backtracking,meng2018exploiting,SCoRF,ValentinNSFIT15,Brachmann_2017_CVPR,Brachmann_2018_CVPR,Brachmann_2019_ICCV,Brachmann_2019_ICCV_NG,brachmann2020visual,Budvytis2019,bui2018scene,Cavallari_corr_19,Li2018,Li_Ylioinas_Verbeek_Kannala_2018,Massiceti2017,li2020hierarchical,Yang_2019_ICCV,zhou2020kfnet}, unlike pose regression methods, focus on predicting 2D-3D correspondences directly from the image, and the camera pose can be solved using the predicted correspondences as in the conventional pipeline.

\noindent \textbf{Continual learning.} Regularization methods penalize changes in weight parameters important for previous tasks~\cite{ aljundi_2018,kirkpatrick_2016,nguyen_2018}. Modular approaches~\cite{aljundi_2017,fernando_2017,chang_2018,rosenbaum_2018} assign new task-specific parameters for each new task amounting to zero forgetting. However, this comes at additional memory requirements. Meta-learning based approaches~\cite{prabhu_2020, javed_2019, beaulieu_2019} use meta-learning to learn sequential tasks. Replay based methods~\cite{rebuffi_2017, chrysakis2020online, iscen_2020} use knowledge distillation~\cite{hinton_2015} to rehearse using a small episodic memory of data stored from previous tasks. On the other hand, several works~\cite{lopez_2017, chaudhry_2019, aljundi_2019} use the episodic memory as an optimization constraint that penalizes increase in loss at previous tasks.

Reservoir sampling~\cite{vitter1985random} samples a subset of data. Aljundi \etal~\cite{aljundi_2019} proposes two sampling methods that attempt to maximize the gradient directions of the stored samples in buffer memory. Recently proposed sampling method by Chrysakis \etal~\cite{chrysakis2020online} propose a balanced buffering strategy to deal with imbalanced class distribution.

\vspace{10px}
\section{Methods}
\label{sec:methods}

In this section, we provide a brief background of continual learning, and the visual localization pipeline used in the continual learning setting. This is followed by the proposed buffering strategy to adapt existing continual learning solutions for visual localization problem.
\vspace{5px}
\subsection{Continual Learning}
\vspace{5px}
We consider a parameterized mapping such as deep-neural network $f \hspace{2mm} o \hspace{2mm} g: x \rightarrow \hat{y}$, where $x \in \mathbb{R}^{W \times H \times 3}$ is the input image, $g_\Theta: x \rightarrow \tilde{y}$ encodes the intermediate representation and $f_\theta: \tilde{y} \rightarrow \hat{y}$ maps the intermediate representation to the final output space.

Given a stream of non-stationary iid data, $D_t$, $t=1...T$, continual learning aims to learn the parameters of the model $f \hspace{2mm} o \hspace{2mm} g$ using the following loss function:

\begin{equation}
    L =   \argmin_{\theta, \Theta} \sum_{t=1}^T{L_t}
\label{eq:cont_loss}
\end{equation}

\noindent where $L_t = \mathbb{E}_{(x, y) \sim D_t} e(\hat{y}, y)$, $e(.)$ refers to the error function such as Euclidean Loss or Cross-Entropy loss between $\hat{y}$ and corresponding ground-truth labels, $y$.

\noindent \textbf{Buffering.} To prevent catastrophic forgetting, a small amount of previous data is stored in a buffer of fixed size, $B$. Input images from the current task/class and corresponding labels are stored in the buffer. We refer to this process of storing images in the buffer as \textit{Img-buff}.

Apart from images, intermediate representations are also stored that provide a better manifold structure. For example, storing pre-softmax layer logits provides a distribution of class probabilities that encodes inter-class semantic relationships. The buffer $B_z$ stores these representations, $\tilde{y}$ w.r.t each image in the buffer, $B$. Buffering intermediate representations is referred to as \textit{Rep-buff}.
    
\noindent \textbf{Replay.} Replay is the process of re-iterating through samples from past scenes stored in the buffer while learning the current task. The final loss is computed for both the current task samples and those from buffer, $B$ as:

\begin{equation}
L = L_t + \mathbb{E}_{(x, y) \sim B} e(\hat{y}, y)
\label{eq:img_only}
\end{equation}

The intermediate representations stored in $B_z$ can be used as pseudo-labels through the process of knowledge distillation. For example, logits from the current network state are constrained to be similar to corresponding ones stored in buffer memory, $B_z$.

\begin{equation}
L = L_t + \mathbb{E}_{(x, y, \tilde{y}) \sim B_z} (e(\hat{y}, y) + e(\hat{\tilde{y}}, \tilde{y}))
\label{eq:rep_only}
\end{equation}    

\noindent \textbf{Algorithms.}
    The buffering algorithm decides which samples in the current task are to be stored for future replay and which samples stored in the buffer are to be replaced. The first stage consists of filling the buffer until it is full. The second stage then decides buffering probability of additional incoming instances. Here we discuss two baseline approaches:
    
    \noindent \textit{Reservoir} sampling assigns the buffering probability of a new instance as $|B|/N$, where $N$ is the total number of instances observed.
    
    \noindent \textit{Class-balance}: One of the limitations with reservoir sampling is class-imbalanced problem -- cardinality difference between different classes in the buffer. As a result, there is also an imbalance in replay rate of different class instances. To create a balanced buffer, only the instances from the largest class are replaced once the buffer is full. If the class $c$ corresponding to current sample $(x)$ itself is the largest class, then one of its sample in the buffer is replaced with probability $m_c/n_c$ where $m_c$ is the number of currently stored instances of $c$, while $n_c$ is the total instances observed of class $c$.
            
The pseudo codes for \textit{Reservoir} and \textit{Class-balance} sampling are presented in the supplementary material. 

\subsection{Visual Localization}
Visual localization aims to estimate the 6DoF camera pose from a given input image. While many deep learning methods have been proposed, we focus on a particular class of methods: deep structured models that have shown to be more accurate than feature based methods~\cite{Brachmann_2018_CVPR,Brachmann_2019_ICCV,li2020hierarchical}. Among these methods, only a recently proposed \textit{HSCNet}~\cite{li2020hierarchical} has shown scalability to large number of disjoint scenes with a single deep model. Unlike related works, \textit{HSCNet}~\cite{li2020hierarchical} maintains an implicit representation of the scene in a set of parameterized hierarchical network layers that predict 3D scene coordinates for each 2D pixel location. Using PnP, the 2D-3D correspondences are used to obtain the final query camera estimate.

 The ground-truth 3D points, $y_{3D}$ are hierarchically clustered into coarse-to-fine set of discrete labels, $y_{l}$, where $l=1,2...L$ are the coarse-to-fine cluster levels such that $|y_1| < |y_2| ... < |y_L| < |y_{3D}|$ . The combined label set is defined as $\mathbf{y} = y_{3D} \bigcup \{y_l\}_{l=1}^L$ with $\mathbf{y}(x)$ corresponding to the labels for input $x$. We refer the reader to the supplementary for the detailed explanation. For each input pixel in a given image, \textit{HSCNet}~\cite{li2020hierarchical} predicts the corresponding cluster label in coarse-to-fine manner, where finer predictions are conditioned on coarser predictions from previous layers using conditioning layers. The task loss for visual localization of input image $x$ is the sum of losses at each cluster level summed over all pixels:
 
\begin{equation}
    L_{t} = \sum_{l=1}^L{\alpha_l \cdot e(\hat{y}_l, y_l(x))} + \beta \cdot e(\hat{y}_{3D}, y_{3D}(x))
\label{eq:visual_task_loss}
\end{equation}
\noindent where $\alpha_l$ and $\beta$ are the weighting coefficients.

In a continual learning setup, the scenes are presented sequentially, and the continual loss function in Eq.~\ref{eq:cont_loss} is computed using Eq.~\ref{eq:visual_task_loss}. For \textit{Img-buff} only the input images and corresponding 3D scene coordinates, $\mathbf{y}$ are stored in $B$. In addition, the intermediate cluster-level predictions, $\mathbf{\tilde{y}} = \hat{y}_{3D} \bigcup \{\hat{y}_l\}_{l=1}^L$ are also stored for \textit{Rep-buff} (\cf Sec.~\ref{sec:exp}).

\begin{algorithm}[t]
\caption{Buffering}
\ \ 
\label{algo:buffering}
\begin{algorithmic}[1] 
\STATE{input stream: ${(x,\mathbf{y}} \sim D_t)$}
\STATE{$c \equiv $ class(x)}
\STATE{Buffer Memory: $B$}
\FOR{$i = 1$ to $n$}
    \IF{$B$ is not \textbf{filled}}
        \STATE{$B \leftarrow (x, y)$}
    
    \ELSE{}
        \STATE{Reservoir / Class-balance / Buff-CS}

    \ENDIF
\ENDFOR
\end{algorithmic}
\end{algorithm}

\begin{algorithm}[t]
\caption{Buff-CS}
\label{algo:buff-cs}
\begin{algorithmic}[1] 
\IF{$c$ is not largest}
    \STATE{Select a random instance of largest class}
    \STATE{Replace it with $(x, \mathbf{y})$}
\ELSE
    \STATE{Flag $\leftarrow \emptyset$ }
    \STATE{cs $\leftarrow$ $CoverageScore$(c, ({$x, \mathbf{y}$}))}
    \IF{ $cs$ is not $\emptyset$}
        \STATE{Flag $\leftarrow$ 1}
    \ELSE
        \STATE{$m_c \leftarrow$ number of currently stored instances of $c$}
        \STATE{$n_c \leftarrow$ number of total instances observed of $c$}
        \STATE{u $\sim$ Random(0,1)}    
        \IF{$u < m_c/n_c$}
            \STATE{Flag $\leftarrow$ 1}
        \ENDIF
    \ENDIF
    \IF{Flag is not $\emptyset$}
    \STATE{Replace an instance of $c$ with ($x, \mathbf{y}$)}
    \ELSE
    \STATE{Ignore}
\ENDIF
\ENDIF

\end{algorithmic}

\end{algorithm}

\begin{algorithm}[t]

\caption{$CoverageScore$}
\textbf{input:} class c  \\
\textbf{input:} new instance $(x, \mathbf{y})$
\begin{algorithmic}[1]
\STATE{Sample data $\{x_b,\mathbf{y}_b\}_{b=1}^{|B_c|}$ of class $c$ from $B$}
\STATE{Compute $cs_1$ using Eq. \ref{eq:cgf_1} }
\RETURN{$cs_1$}
\end{algorithmic}
\label{algo:coverageS}
\end{algorithm}

\subsection{Coverage Score Buffering}
Unlike classification problems, visual localization scenes/classes are visually diverse and independent - learning localization on images of a particular sub-scene does not enable generalization to other parts of the scene. To retain localization performance on all sub-scenes of a given scene, the buffer needs to maintain images that maximize the scene coverage. In this section we propose a method to sample images that provides an improved scene coverage - referred to as \textit{Buff-CS}.

The fundamental difference with \textit{Class-balance} is in the case where the current scene is the largest class in the buffer, and each new instance is buffered with probability $m_c/n_c$. With increasing $n_c$, or for small $m_c$ the buffering probability decreases sharply. Thereby, later observations have a lower probability of being buffered\footnote{Note that when current scene is not largest, the sample will have buffering probability 1}. In this work, we increase the buffering probability to 1 if the incoming new instance provides novel scene observations compared to the instances that observed by the buffer images. Example scene observations can be the ground-truth 3D scene coordinates. Given the 3D scene coordinates $y_{3D}(x)$ seen by the new instance $x$ and those by buffer images of class $c, y_{3D}(B_c) = \{y_{3D}(x_b)\}_{b=1}^{|B_c|}$ we compute the coverage score factor:

\begin{equation}
    cs_{3D} = y_{3D}(x) \setminus y_{3D}(B_c)
    \label{eq:cgf_3d}
\end{equation}

\noindent If $cs_{3D}$ is not $\emptyset$, $x$ observes novel 3D points that are not seen by the images $I \in B_c$ and hence will be registered into the buffer, $B$.

As the coverage score is computed for each sample during the buffering process, an efficient method is required to compute the coverage score. Although $y_{3D}$ provides an accurate estimate of coverage score, for dense 3D models this becomes computationally intensive. We propose an efficient method to compute the coverage score by replacing $y_{3D}$ in Eq. \ref{eq:cgf_3d} with the coarsest cluster level $y_1$:

\begin{equation}
    cs_{1} = y_{1}(x) \setminus y_{1}(B_c)
    \label{eq:cgf_1}
\end{equation}

\noindent As $|y_1| \ll |y_{3D}|$ the computational efficiency is significantly improved at the cost of lower coverage score accuracy. The gain in coverage score obtained using coarse-level cluster information (c.f. Table~\ref{tab:all_results}) indicates that the approximate method still achieves higher coverage score than \textit{Class-balance} method. Algorithms ~\ref{algo:buff-cs}
 and ~\ref{algo:coverageS} are the pseudo codes for \textit{Buff-CS} and our coverage score method.

\section{Experiments}
\label{sec:exp}

In this section, we describe our experimental setup.
\subsection{Benchmarks and Environment Setup}
\noindent \textbf{Datasets.}
Different from standard \textit{continual learning} tasks evaluated on classification benchmarks, we select two \textit{visual localization} benchmarks, \textbf{7Scenes}~\cite{SCoRF} and \textbf{12Scenes}~\cite{valentin2016learning} for the experiments. \textbf{7Scenes} records RGB-D image sequences of seven different indoor scenes from a handheld Kinect camera, each sequence consists of 500--1000 frames at $640\times480$ resolution. \textbf{12Scenes} is another indoor RGB-D dataset containing 4 large scenes, with a total of 12 rooms, captured using a Structure.io depth sensor coupled with an iPad color camera. Both datasets also provide dense 3D points and the ground truth camera poses. In order to evaluate the \textbf{CL} methods in a sequential manner, we integrate the individual seven scenes and twelve scenes into single coordinate systems and yields to two large scenes similar to~\cite{Brachmann_2019_ICCV,li2020hierarchical}. In addition, we also synthesize the largest scene by the combination of all nineteen scenes. These three large scenes are denoted by \textbf{i7S} (ca. $125m^3$ total), \textbf{i12S} (ca. $520m^3$ total), and  \textbf{i19S} (ca. $645m^3$ total), respectively.

\noindent \textbf{Baselines.} In this work, we adopt two buffering methods as the baselines,  namely \textit{Reservoir}~\cite{vitter1985random} and \textit{ Class-balance}~\cite{chrysakis2020online}. These two methods are referred to as \textit{Reservoir} and \textit{Class-balance} respectively. \textit{Reservoir} aims to sample $k$ data instances from an input stream of unknown size, where $k$ is the predefined sample size. The data from the input stream in our experiments are the training frames of \textbf{i7S},  \textbf{i12S},  or \textbf{i19S}, and this method guarantees the same probability for the individual frame to be selected into the buffer. \textit{Class-balance} aims to further solve the class-imbalance problem in online continual learning. This method keeps the classes as balanced as possible, while the distribution of each class/scene is preserved.

Besides the aforementioned methods, we also consider one weak baseline --  train our models without buffering and image-replay. We refer to this method as \textit{W/O Buffering}.

\noindent \textbf{Evaluation Metrics.} Following~\cite{SCoRF}, we evaluate the performance of these methods using pose accuracy. Pose accuracy is defined as the percentage of the query images with an error below $5$ cm and $5^\circ$. We consider both the accuracy after the training is complete and the average accuracy over different stages of the training process. Similar to ~\cite{chaudhry2020continual}, the latter one is defined as:

\begin{equation}
     A_{i} = \frac{1}{N-i+1}\sum_{j=i}^N a_{i,j}   
\end{equation}
where $N$ is the total number of scenes, and $a_{i,j}$ denotes the accuracy of the model on scene $i$ after the training of the model on scene $j$ is complete.
\subsection{Implementation Details}

In the task of continual learning for visual localization, individual scenes are fed to the training network in an incremental manner -- that is to say, data in the first scene is trained to estimate scene coordinates, then the training weights are utilized as the initialization for the second scene.

For training HSCNet~\cite{li2020hierarchical} in a continual learning setup, the training data of each scene are sampled and stored in the buffer after the training of the corresponding scene is complete. As mentioned before, buffering only the input images and corresponding labels is referred to as \textit{Img-buff}, and buffering additionally the intermediate representations is referred to as \textit{Rep-buff}. For \textit{Img-buff}, we store the RGB images, the depth maps, and ground truth poses to the buffer, and the ground truth labels for training are generated in the same way as in ~\cite{li2020hierarchical}. For \textit{Rep-buff}, we additionally store pre-softmax layer logits and the scene coordinates predicted by the current model.

We keep most of the training settings in~\cite{li2020hierarchical} unchanged.However, some changes are made to adapt it to our continual learning setup. First, 
all the networks are trained using the Adam~\cite{kingma2014adam} optimizer with a smaller learning rate of $5e^{-5}$. Second, different from Li \etal~\cite{li2020hierarchical} who train each individual scene with 300K iterations, we reduce it to 30K for each scene to save the training time. This is because of the sequential property of the continual learning task, \ie the training on each scene begins only after the previous one is finished. It is reported in~\cite{li2020hierarchical} that the training time for each scene is around 12 hours, and thus the training would become impractical if we keep the number of iterations unchanged for our experiments. Besides, we also found that training with 30K iterations still leads to comparable results.

\section{Results}
\label{sec:results}

\begin{table*}[t!]
\renewcommand\arraystretch{1.1}
\begin{adjustbox}{max width=1.01\textwidth}
\normalsize
\begin{tabular}{|c|l|ccc|ccc|ccc|}
\hline
 &
  \multicolumn{1}{c|}{} &
  \multicolumn{3}{c|}{\textbf{i7S}} &
  \multicolumn{3}{c|}{\textbf{i12S}} &
  \multicolumn{3}{c|}{\textbf{i19S}} \\ \cline{3-11} 
 &
  \multicolumn{1}{c|}{} &
  \multicolumn{1}{c|}{} &
  \multicolumn{2}{c|}{\textbf{ Accuracy ( \% )}} &
  \multicolumn{1}{c|}{} &
  \multicolumn{2}{c|}{\textbf{ Accuracy ( \% )}} &
  \multicolumn{1}{c|}{} &
  \multicolumn{2}{c|}{\textbf{ Accuracy (\% )}} \\ \cline{4-5} \cline{7-8} \cline{10-11} 
\multirow{-3}{*}{\textbf{Buffer Size}} &
  \multicolumn{1}{c|}{\multirow{-3}{*}{\textbf{Buffer methods}}} &
  \multicolumn{1}{c|}{\multirow{-2}{*}{\textbf{\begin{tabular}[c]{@{}c@{}}Coverage Score \\  (average \%)\end{tabular}}}} &
  \multicolumn{1}{l|}{\textbf{Img-buff}} &
  \multicolumn{1}{l|}{\textbf{Rep-buff}} &
  \multicolumn{1}{c|}{\multirow{-2}{*}{\textbf{\begin{tabular}[c]{@{}c@{}}Coverage Score \\  (average \%)\end{tabular}}}} &
  \multicolumn{1}{l|}{\textbf{Img-buff}} &
  \multicolumn{1}{l|}{\textbf{Rep-buff}} &
  \multicolumn{1}{c|}{\multirow{-2}{*}{\textbf{\begin{tabular}[c]{@{}c@{}}Coverage Score \\  (average \%)\end{tabular}}}} &
  \multicolumn{1}{l|}{\textbf{Img-buff}} &
  \multicolumn{1}{l|}{\textbf{Rep-buff}} \\ \hline
 &
  Reservoir &
  72.47 &
  56.8 &
  59.79 &
  58.3 &
  67.51 &
  70.23 &
  39.6 &
  33.2 &
  34.37 \\ \cline{2-2}
 &
  Class-balance &
  78.7 &
  59.6 &
  {\color{red} 64.25} &
  63.9 &
  74.4 &
  75.72 &
  55.2 &
  {\color{red} 47.98} &
  {\color{red} 54.12} \\ \cline{2-2}
\multirow{-3}{*}{128} &
  Buff-CS (ours) &
  {\color{red} 90.11} &
  {\color{red} 61.2} &
  61.13 &
  {\color{red} 66.9} &
  {\color{red} 75.4} &
  {\color{red} 77.33} &
  {\color{red} 58.8} &
  46.43 &
  51.86 \\ \hline
 &
  Reservoir &
  79.3 &
  69.3 &
  69.46 &
  75.4 &
  82.54 &
  85.82 &
  58.6 &
  47.53 &
  49.48 \\ \cline{2-2}
 &
  Class-balance &
  87.3 &
  70.06 &
  69.76 &
  76.5 &
  85.72 &
  87.36 &
  71.1 &
  64.7 &
  67.31 \\ \cline{2-2}
\multirow{-3}{*}{256} &
  Buff-CS (ours) &
  {\color{red} 92.6} &
  {\color{red} 72} &
  {\color{red} 71.34} &
  {\color{red} 86.3} &
  {\color{red} 91.85} &
  {\color{red} 92.78} &
  {\color{red} 76.7} &
  {\color{red} 68} &
  {\color{red} 70.09} \\ \hline
 &
  Reservoir &
  91.3 &
  73.4 &
  72.34 &
  89 &
  93.67 &
  94.33 &
  70.5 &
  60.15 &
  62.34 \\ \cline{2-2}
 &
  Class-balance &
  92.5 &
  74.24 &
  73.6 &
  90.7 &
  95.63 &
  95.63 &
  86.1 &
  78.61 &
  79.06 \\ \cline{2-2}
\multirow{-3}{*}{512} &
  Buff-CS (ours) &
  {\color{red} 97.4} &
  {\color{red} 75.81} &
  {\color{red} 76.06} &
  {\color{red} 95.7} &
  {\color{red} 96.42} &
  {\color{red} 95.85} &
  {\color{red} 91.3} &
  {\color{red} 80.93} &
  {\color{red} 79.51} \\ \hline
 &
  Reservoir &
  95.8 &
  75.7 &
  {\color{red} 77.07} &
  94.2 &
  97.26 &
  97.14 &
  86.3 &
  80.94 &
  79.5 \\ \cline{2-2}
 &
  Class-balance &
  96.9 &
  {\color{red} 77.09} &
  74.71 &
  96.3 &
  98.43 &
  {\color{red} 98.49} &
  93.1 &
  {\color{red} 85.36} &
  84.17 \\ \cline{2-2}
\multirow{-3}{*}{1024} &
  Buff-CS (ours) &
  {\color{red} 98.7} &
  76.89 &
  75.22 &
  {\color{red} 97.7} &
  {\color{red}98.9} &
  98.11 &
  {\color{red} 96} &
  85.23 &
  {\color{red} 85.42} \\ \hline
  
   \multicolumn{2}{|c|}{HSCNet (joint training)~\cite{li2020hierarchical}} & \color{blue} 100 & \multicolumn{2}{c|} {\color{blue} 84.19} & \color{blue} 100 & \multicolumn{2}{c|} {\color{blue} 99.0} & \color{blue} 100 & \multicolumn{2}{c|} {\color{blue} 92.5}\\ \hline

\end{tabular}
\end{adjustbox}
\vspace{3px}
\caption{Coverage score and  accuracy of our method and the two baselines on \textbf{i7S}, \textbf{i12S}, and \textbf{i19S}  after  the  training  is  complete. The coverage scores are averaged across all the scenes. The best and second best results among approaches are highlighted in blue and red respectively. }
\label{tab:all_results}
\end{table*}

\subsection{Comparison with Baselines}

In this section, we compare our \textit{Buff-CS} method against the two strong baselines ( \textit{Reservoir}, and \textit{Class-balance}) on the three combined scenes. 

\tab{all_results} reports the performance in terms of pose accuracy averaged over all the scenes after the training is complete, and the coverage score (Eq. \ref{eq:cgf_1}). Four different buffer sizes are experimented, ranging from $B = 128$ to $B = 1024$.  We also present results for the methods with the two types of buffering information, namely \textit{Img-buff} and \textit{Rep-buff}. 
In addition to the above methods, we report the results of HSCNet in the last row. It is also worth noting that we do not report the 95\% confidence interval for the results in \tab{all_results} as in~\cite{chrysakis2020online,NEURIPS2020_b704ea2c,chaudhry2020continual}, due to the impractical long training time of our task. However, to support our results, we report the 95 \% confidence interval of the accuracy for the results on  \textbf{i7S} in the ablation study (see \tab{95 confidence interval}).

We observe that \textit{Buff-CS} achieves the highest coverage score in all settings and outperforms the other two approaches in most of the experiments in terms of the pose accuracy. With the buffer size $B = 256$ and $B = 512$, the higher coverage score of our method yields better pose accuracy on all the three combined scenes. The results indicate that the performance of the \textit{Reservoir} baseline is significantly exceeded by the other two methods due to the low coverage score. We notice that the \textit{Class-balance} method achieves performance comparable to \textit{Buff-CS} with the buffer size $B = 128$ and $B = 1024$. We see that with a large buffer size $B = 1024$, the coverage score for these two approaches are both over 93\%, which indicates that nearly all parts of the scene appear in the buffer and the effectiveness of using coverage score is narrowed. Other factors such as the robustness of RANSAC~\cite{RANSAC} in camera pose estimation also affect the final results. With extremely small buffer size $B = 128$ on \textbf{i19S}, both \textit{Class-balance} and \textit{Buff-CS} on an average perform comparably and better than \textit{Reservoir}. When comparing the two buffering strategies \textit{Img-buff} and \textit{Rep-buff} we observe that \textit{Rep-buff} performs better on larger scenes and smaller buffer length. In particular, the biggest performance gap between \textit{Img-buff} and \textit{Rep-buff} is observed in \textbf{i19S} and $B = 128$. For larger buffer size, $B = 1024$ the performance is comparable. A detailed analysis of \textit{Rep-buff} is provided in the supplementary. Although the performance of CL approaches lags behind the joint training setting (last row in Table~\ref{tab:all_results}), memory (\cf Sec.~\ref{sec:mem}) and computational efficiency provides sufficient motivation to pursue visual localization in CL setting.

\begin{table}[t!]
\renewcommand\arraystretch{1.4}
\resizebox{.48\textwidth}{!}{
\normalsize
\begin{tabular}{lccccc}
\hline
                 & \multicolumn{5}{c}{\textbf{Accuracy (\%)}}                                                                                       \\
\multirow{-2}{*}{\textbf{Scene}} & \textbf{W/O Buffering} & \textbf{Revervoir} & \textbf{Class-balance} & \textbf{Buff-CS( ours)} & \multicolumn{1}{l}{\textbf{HSCNet}~\cite{li2020hierarchical}} \\ \hline
Chess            & 0.0   & {\color{red} 91.45} & 88.50                        & 88.30                         & {\color{blue} 97.30}  \\
Fire             & 0.0   & 67.25                        & 76.35                       & {\color{red} 80.30}  & {\color{blue} 96.15} \\
Heads            & 0.0   & 69.40                         & 80.40                        & {\color{red} 89.60}  & {\color{blue} 98.30}  \\
Office           & 0.0   & {\color{red} 71.00}    & 62.98                       & 62.08                        & {\color{blue} 85.50}  \\
Pumpkin          & 0.0   & 50.60                         & 51.45                       & {\color{red} 54.25} & {\color{blue} 60.85} \\
Kitchen          & 0.0   & {\color{red} 56.14} & 50.18                       & 49.78                        & {\color{blue} 63.74} \\
Stairs           & 74.30  & 79.70                         & {\color{red} 80.60} & 79.70                         & {\color{blue} 87.50}  \\ \hline
\textbf{Average} & 10.61 & 69.30                         & 70.06                       & {\color{red} 72.00}    & {\color{blue} 84.19} \\ \hline
\end{tabular}
}
\vspace{5px}
\caption{The percentage of accurately localized test images (error $<$ 5 cm, $5^\circ$) on \textbf{i7S} with the buffer size $B = 256$,  after  the  training  is  complete. Here we use \textit{Img-buff} for replay. The best and second best results are highlighted in blue and red respectively. } 
\label{tab:7S_results}

\end{table}

\begin{figure}[t!]
  \centering
        \includegraphics[width=\linewidth]{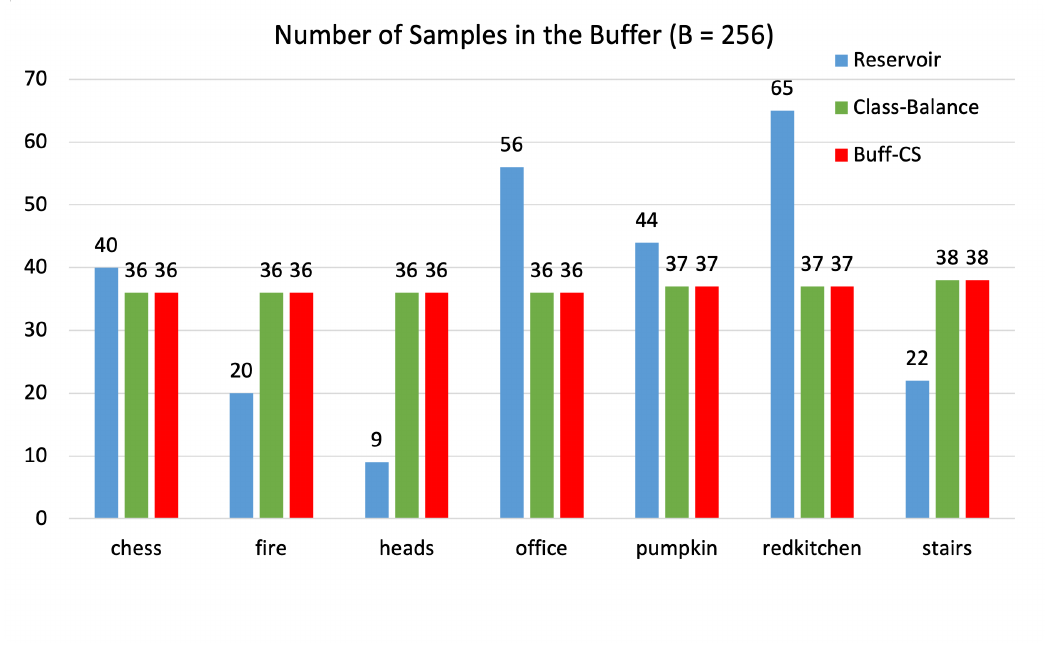}
        \vspace{-20px}
        \caption{The memory distribution of three methods with buffer size = 256 after the training is complete on \textbf{i7S}. \textit{Reservoir} suffers from data imbalance while the samples in \textit{Class-balance} and \textit{Buff-CS} are balanced.
        }
        \label{fig:num samples}
\end{figure}

\begin{figure*}
\begin{center}

   \includegraphics[width=0.9\linewidth]{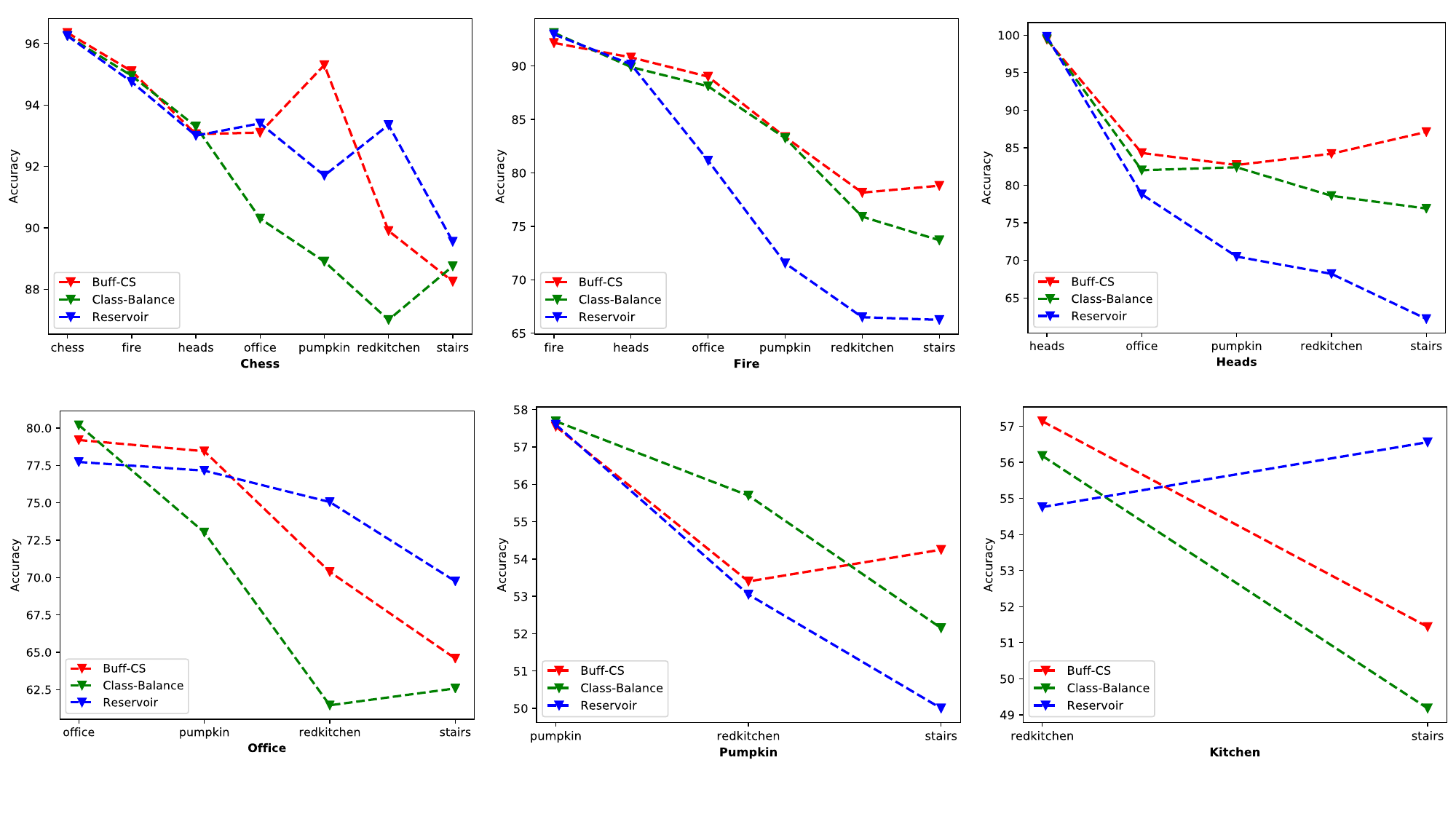}
   \end{center}
   \vspace{-20px}
   \caption{The accuracy (error $< 5$ cm, $5^\circ$) on individual scenes of \textbf{i7S} (except for the last scene) at each stage of the training. The x axis indicates the training progress. All methods employs an \textit{Img-buff} buffer of size 256.}
        \label{fig:accuracy}
\vspace{-5px}
\end{figure*}

To make a more detailed comparison among the approaches, we report the accuracy on each scene of i7S after training is complete.
 ~\tab{7S_results} presents the results for the methods with \textit{Img-buff} and buffer size $B = 256$.
Indeed, training on all scenes together as in~\cite{li2020hierarchical} achieves the best performance. When we train the model in the incremental scenario without buffering, as shown in \tab{7S_results}, the accuracy on the previously encountered scenes is $0 \%$, which indicates that the visual localization network also suffers from catastrophic forgetting when trained in a continual manner. \textit{Reservoir} exceeds both \textit{Class-balance} and \textit{Buff-CS} methods on Chess, Office, and Redkitchen. This can be attributed to the larger number of samples stored in the buffer corresponding to these scenes, see \fig{num samples} for the sample distribution. However, the accuracy drops dramatically on Fire and Heads due to the reduced number of scene samples in the buffer. \textit{Buff-CS} manages to balance the number of samples in the buffer and effectively improves the coverage score compared to \textit{Class-balance}. Thus, it achieves generally better accuracy among all the sampling approaches while maintaining a balanced class distribution in the buffer.

\begin{table}[t!]
\renewcommand\arraystretch{1.2}
\hspace{2mm}
\resizebox{.45\textwidth}{!}{
\normalsize
\begin{tabular}{lccc}
\hline
\multirow{2}{*}{\textbf{Scene}} & \multicolumn{3}{c}{\textbf{Average Accuracy (\%)}}\\
& \textbf{Reservoir} & \textbf{Class-balance} & \textbf{Buff-CS (ours)}             \\ \hline
Chess   & {\color[HTML]{FE0000} 93.14} & 91.35                        & 93.01                        \\
Fire    & 78.09                        & 83.99                        & {\color[HTML]{FE0000} 85.38} \\
Heads   & 75.90                         & 83.88                        & {\color[HTML]{FE0000} 87.54} \\
Office  & {\color[HTML]{FE0000} 74.92} & 69.31                        & 73.15                        \\
Pumpkin & 53.55                        & {\color[HTML]{FE0000} 55.18} & 55.07                        \\
Kitchen & {\color[HTML]{FE0000} 55.66} & 52.68                        & 54.29                        \\
Stairs  & 78.90                         & {\color[HTML]{FE0000} 79.40}  & 79.30                         \\ \hline
\textbf{Overall Average}    & 72.88              & 73.68                  & {\color[HTML]{FE0000} 75.39} \\ \hline
\end{tabular}
}
\vspace{10px}
\caption{The average accuracy over different stages of the training process on each scene of \textbf{i7S} with the buffer size $B = 256$. Our method has overall better performance compared to the other two methods.} 
\label{tab:average accuracy}

\end{table}

The average accuracy in  \tab{average accuracy} evaluates the performance of the three methods on previous tasks after completing a new task. \tab{average accuracy} presents the average accuracy for each scene in \textbf{i7S} with buffer size $B = 256$. Similar to \tab{7S_results}, \textit{Reservoir} achieves best performance on Chess, office, and Redkitchen while it drops significantly on Fire, Heads (around 10\%) due to the class imbalance. In terms of \textit{overall average}, this method falls behind \textit{Class-balance} and \textit{Buff-CS} by 0.8\% and 2.51\% respectively. \textit{Class-balance} relieves the problem by balancing the sample distribution. However, it still has weaker results than \textit{Buff-CS} due to the lower coverage score. \fig{accuracy} shows a more detailed picture in terms of the test accuracy of the three methods after each task is completed. First, we observe that with increasing task length, the performance generally drops across all methods and scenes. This is due to the decreasing of class samples in the buffer. Second, compare to \textit{Class-balance}, \textit{Buff-CS} shows strong performance in the majority of cases. In such a scenario, we believe that the increase of coverage score has a positive effect on test accuracy by providing larger scene observations during the replay process.

\subsection{Ablation Study}
In this section, we  conduct an ablation study to illustrate how different factors affect the performance of the localization system in the continual learning setup. We conduct the experiments on \textbf{i7S} with buffer size $B = 256$ and \textit{Img-buff} information.

\noindent \textbf{Disorder Scenes.} We generate random permutations of the scene order being fed to the training network in a continual manner. Results presented in Table 1 of the supplementary shows that the \textit{Buff-CS} performs comparably or better than the baseline methods.

\noindent \textbf{95\% confidence interval.} We experience intractable training time when trying to report the 95\% confidence interval for all of the experiments in \tab{all_results}. Thus, only 95\% confidence interval of the test set accuracy on \textbf{i7S} with buffer size 256 is reported in \tab{95 confidence interval}. The experiments are run 5 times with different random seeds, and we keep the same seed for all approaches in each run. We observe that the conclusions in \tab{all_results} still holds, \ie 
our method outperforms the two baselines with both \textit{Img-buff} and \textit{Rep-buff}.

\begin{table}[t!]
\renewcommand\arraystretch{1.2}
\hspace{2mm}
\resizebox{.45\textwidth}{!}{
\normalsize

\begin{tabular}{cccc}
\hline
                                  & \multicolumn{3}{c}{\textbf{Final Accuracy (\%)}}                                   \\
\multirow{-2}{*}{\textbf{Buff Size = 256}} & \textbf{Reservoir}  & \textbf{Class-balance} & \textbf{Buff-CS (ours)}                    \\ \hline
Img-buff                          & $66.99 \pm 1.23$ & $69.66 \pm 0.54$    & {\color{red} $70.51 \pm 0.94$} \\
Rep-buff                  & $68.00 \pm 1.32$    & $69.12 \pm 1.67$    & {\color{red} $70.67 \pm 1.10$}  \\ \hline
\end{tabular}
}
\vspace{10px}
\caption{95\% confidence interval of the average accuracy on \textbf{i7S} with $B = 256$ after  the  training  is  complete. The results are obtained over 5 runs.}
\label{tab:95 confidence interval}
\vspace{-5px}
\end{table}

\subsection{Training Time and Memory Consumption}
\label{sec:mem}
Visual localization in continual learning setup aims to achieve data efficiency compared to joint training by learning the tasks sequentially. However, due to catastrophic forgetting, the concept of replay-buffer is used which incurs memory costs of its own. In this section we analyze the memory requirements of buffering different forms of data and compare to the data storage costs of training jointly on all tasks.

To guarantee a fair comparison, all of the experiments are run on NVIDIA Tesla V100 GPUs. We observe that, with buffer size $B = 256$, \textit{Reservoir}, \textit{Class-balance}, and \textit{Buff-CS} require roughly the same amount of time ($\sim 20 h$) with both \textit{Img-buff} and \textit{Rep-buff}, which is reasonable since these methods have similar buffering and replay process. When comparing the efficiency between \textit{Img-buff} and \textit{Rep-buff}, we 
observe that \textit{Rep-buff} is more memory-consuming. Approximately 10 times the more storage space ($\sim 1091$ Mb) is needed compared to \textit{Img-buff} ($\sim 117$ Mb), as it requires larger space for storing dense intermediate cluster predictions.

Compared to the joint training setting which requires to store $\sim 35$ Gb for \textbf{i19S}, the proposed CL method only requires on an average $1.9 $ Gb and $2.8$ Gb with \textit{Img-buff} and \textit{Rep-buff} buffering respectively. From this occupied space, $1.8$ Gb corresponds to the average space for task-specific data, while the remaining is allotted to buffer data.

\section{Conclusion}
\label{sec:concl}

In this work we have presented the problem of continual visual localization. A strong baseline is introduced based on experience replay using samples from a small fixed-size buffer. This prevents catastrophic forgetting while learning localization on new scenes. We propose a new buffering strategy that takes into account the 3D scene geometry while keeping a balanced distribution of class samples. The proposed method is evaluated on several indoor localization datasets demonstrating better or competitive performance against the baselines across various settings. Instead of single scene per task, multiple scenes can be considered which makes the problem more challenging and a direction for future work. Although the proposed method balances inter-task data distributions, the above problem setting also requires balancing intra-task data from multiple scenes.

\noindent \textbf{Acknowledgements.} This  work  has  been  supported  by  the  Academy of  Finland  (grants 309902, 327911). The authors would like to acknowledge the computational resources provided by the Aalto Science-IT project and CSC-IT Center for Science, Finland.

{\small
\bibliographystyle{ieee_fullname}
\bibliography{egbib}

\begin{thebibliography}{10}\itemsep=-1pt

\bibitem{aljundi_2018}
Rahaf Aljundi, Francesca Babiloni, Mohamed Elhoseiny, Marcus Rohrbach, and
  Tinne Tuytelaars.
\newblock Memory aware synapses: Learning what (not) to forget.
\newblock In {\em ECCV}, 2018.

\bibitem{aljundi_2017}
Rahaf Aljundi, Punarjay Chakravarty, and Tinne Tuytelaars.
\newblock Expert gate: Lifelong learning with a network of experts.
\newblock In {\em CVPR}, 2017.

\bibitem{aljundi_2019}
Rahaf Aljundi, Min Lin, Baptiste Goujaud, and Yoshua Bengio.
\newblock Gradient based sample selection for online continual learning.
\newblock In {\em NeurIPS}, 2019.

\bibitem{Balntas_2018_ECCV}
Vassileios Balntas, Shuda Li, and Victor~Adrian Prisacariu.
\newblock {RelocNet}: Continuous metric learning relocalisation using neural
  nets.
\newblock In {\em ECCV}, 2018.

\bibitem{beaulieu_2019}
Shawn Beaulieu, Lapo Frati, Thomas Miconi, Joel Lehman, Kenneth~O. Stanley,
  Jeff Clune, and Nick Cheney.
\newblock Learning to continually learn.
\newblock In {\em ECAI}, 2020.

\bibitem{Brachmann_2017_CVPR}
Eric Brachmann, Alexander Krull, Sebastian Nowozin, Jamie Shotton, Frank
  Michel, Stefan Gumhold, and Carsten Rother.
\newblock {DSAC} - {D}ifferentiable {RANSAC} for camera localization.
\newblock In {\em CVPR}, 2017.

\bibitem{BrachmannMKYGR16}
Eric Brachmann, Frank Michel, Alexander Krull, Michael~Ying Yang, Stefan
  Gumhold, and Carsten Rother.
\newblock Uncertainty-driven {6D} pose estimation of objects and scenes from a
  single {RGB} image.
\newblock In {\em CVPR}, 2016.

\bibitem{Brachmann_2018_CVPR}
Eric Brachmann and Carsten Rother.
\newblock Learning less is more - {6D} camera localization via {3D} surface
  regression.
\newblock In {\em CVPR}, 2018.

\bibitem{Brachmann_2019_ICCV}
Eric Brachmann and Carsten Rother.
\newblock Expert sample consensus applied to camera re-localization.
\newblock In {\em ICCV}, 2019.

\bibitem{Brachmann_2019_ICCV_NG}
Eric Brachmann and Carsten Rother.
\newblock Neural-guided {RANSAC}: Learning where to sample model hypotheses.
\newblock In {\em ICCV}, 2019.

\bibitem{brachmann2020visual}
Eric Brachmann and Carsten Rother.
\newblock Visual camera re-localization from {RGB} and {RGB-D} images using
  {DSAC}.
\newblock {\em arXiv:2002.12324}, 2020.

\bibitem{mapnet2018}
Samarth Brahmbhatt, Jinwei Gu, Kihwan Kim, James Hays, and Jan Kautz.
\newblock Geometry-aware learning of maps for camera localization.
\newblock In {\em CVPR}, 2018.

\bibitem{Budvytis2019}
Ignas Budvytis, Marvin Teichmann, Tomas Vojir, and Roberto Cipolla.
\newblock Large scale joint semantic re-localisation and scene understanding
  via globally unique instance coordinate regression.
\newblock In {\em BMVC}, 2019.

\bibitem{bui2018scene}
Mai Bui, Shadi Albarqouni, Slobodan Ilic, and Nassir Navab.
\newblock Scene coordinate and correspondence learning for image-based
  localization.
\newblock In {\em BMVC}, 2018.

\bibitem{NEURIPS2020_b704ea2c}
Pietro Buzzega, Matteo Boschini, Angelo Porrello, Davide Abati, and Simone
  Calderara.
\newblock Dark experience for general continual learning: a strong, simple
  baseline.
\newblock In {\em NeurIPS}, 2020.

\bibitem{Cavallari_corr_19}
Tommaso Cavallari, Luca Bertinetto, Jishnu Mukhoti, Philip Torr, and Stuart
  Golodetz.
\newblock Let's take this online: Adapting scene coordinate regression network
  predictions for online {RGB-D} camera relocalisation.
\newblock In {\em 3DV}, 2019.

\bibitem{CavallariGLVST17}
Tommaso Cavallari, Stuart Golodetz, Nicholas~A Lord, Julien Valentin, Luigi
  Di~Stefano, and Philip~HS Torr.
\newblock On-the-fly adaptation of regression forests for online camera
  relocalisation.
\newblock In {\em CVPR}, 2017.

\bibitem{cavallari2019real}
Tommaso Cavallari, Stuart Golodetz, Nicholas~A Lord, Julien Valentin, Victor~A
  Prisacariu, Luigi Di~Stefano, and Philip~HS Torr.
\newblock Real-time rgb-d camera pose estimation in novel scenes using a
  relocalisation cascade.
\newblock {\em PAMI}, 42(10):2465--2477, 2019.

\bibitem{chaudhry2021using}
Arslan Chaudhry, Albert Gordo, Puneet~K. Dokania, Philip Torr, and David
  Lopez-Paz.
\newblock Using hindsight to anchor past knowledge in continual learning.
\newblock In {\em AAAI}, 2021.

\bibitem{chaudhry2020continual}
Arslan Chaudhry, Naeemullah Khan, Puneet Dokania, and Philip Torr.
\newblock Continual learning in low-rank orthogonal subspaces.
\newblock In {\em NeurIPS}, 2020.

\bibitem{chaudhry_2019}
Arslan Chaudhry, Marc'Aurelio Ranzato, Marcus Rohrbach, and Mohamed Elhoseiny.
\newblock Efficient lifelong learning with a-gem.
\newblock In {\em ICLR}, 2018.

\bibitem{chrysakis2020online}
Aristotelis Chrysakis and Marie-Francine Moens.
\newblock Online continual learning from imbalanced data.
\newblock In {\em ICML}, 2020.

\bibitem{Ding_2019_ICCV}
Mingyu Ding, Zhe Wang, Jiankai Sun, Jianping Shi, and Ping Luo.
\newblock {CamNet}: Coarse-to-fine retrieval for camera re-localization.
\newblock In {\em ICCV}, 2019.

\bibitem{fernando_2017}
Chrisantha Fernando, Dylan Banarse, Charles Blundell, Yori Zwols, David Ha,
  Andrei~A Rusu, Alexander Pritzel, and Daan Wierstra.
\newblock Pathnet: Evolution channels gradient descent in super neural
  networks.
\newblock In {\em arXiv preprint arXiv:1701.08734}, 2017.

\bibitem{RANSAC}
Martin~A Fischler and Robert~C Bolles.
\newblock Random sample consensus: A paradigm for model fitting with
  applications to image analysis and automated cartography.
\newblock {\em {CACM}}, 24(6):381--395, 1981.

\bibitem{Guzman-RiveraKGSSFI14}
Abner Guzm{\'{a}}n{-}Rivera, Pushmeet Kohli, Ben Glocker, Jamie Shotton, Toby
  Sharp, Andrew~W. Fitzgibbon, and Shahram Izadi.
\newblock Multi-output learning for camera relocalization.
\newblock In {\em CVPR}, 2014.

\bibitem{hinton_2015}
Geoffrey Hinton, Oriol Vinyals, and Jeff Dean.
\newblock Distilling the knowledge in a neural network.
\newblock In {\em arXiv preprint arXiv:1503.02531}, 2015.

\bibitem{chang_2018}
Yen-Chang Hsu, Yen-Cheng Liu, Anita Ramasamy, and Zsolt Kira.
\newblock Re-evaluating continual learning scenarios: A categorization and case
  for strong baselines.
\newblock In {\em NeurIPS workshops}, 2018.

\bibitem{iscen_2020}
Ahmet Iscen, Jeffrey Zhang, Svetlana Lazebnik, and Cordelia Schmid.
\newblock Memory-efficient incremental learning through feature adaptation.
\newblock In {\em ECCV}, 2020.

\bibitem{javed_2019}
Khurram Javed and Martha White.
\newblock Meta-learning representations for continual learning.
\newblock In {\em NeurIPS}, 2019.

\bibitem{KendallC15bay}
Alex Kendall and Roberto Cipolla.
\newblock Modelling uncertainty in deep learning for camera relocalization.
\newblock In {\em ICRA}, 2016.

\bibitem{Kendall_2017_CVPR}
Alex Kendall and Roberto Cipolla.
\newblock Geometric loss functions for camera pose regression with deep
  learning.
\newblock In {\em CVPR}, 2017.

\bibitem{kendall2015convolutional}
Alex Kendall, Matthew Grimes, and Roberto Cipolla.
\newblock {PoseNet}: A convolutional network for real-time {6-DoF} camera
  relocalization.
\newblock In {\em ICCV}, 2015.

\bibitem{kingma2014adam}
Diederik~P Kingma and Jimmy Ba.
\newblock Adam: A method for stochastic optimization.
\newblock {\em arXiv preprint arXiv:1412.6980}, 2014.

\bibitem{kirkpatrick_2016}
James Kirkpatrick, Razvan Pascanu, Neil Rabinowitz, Joel Veness, Guillaume
  Desjardins, Andrei~A Rusu, Kieran Milan, John Quan, Tiago Ramalho, Agnieszka
  Grabska-Barwinska, et~al.
\newblock Overcoming catastrophic forgetting in neural networks.
\newblock {\em PNAS}, 114(13):3521--3526, 2017.

\bibitem{LaskarMKK17}
Zakaria Laskar, Iaroslav Melekhov, Surya Kalia, and Juho Kannala.
\newblock Camera relocalization by computing pairwise relative poses using
  convolutional neural network.
\newblock In {\em ICCV Workshops}, 2017.

\bibitem{li2020hierarchical}
Xiaotian Li, Shuzhe Wang, Yi Zhao, Jakob Verbeek, and Juho Kannala.
\newblock Hierarchical scene coordinate classification and regression for
  visual localization.
\newblock In {\em CVPR}, 2020.

\bibitem{Li2018}
Xiaotian Li, Juha Ylioinas, and Juho Kannala.
\newblock Full-frame scene coordinate regression for image-based localization.
\newblock In {\em RSS}, 2018.

\bibitem{Li_Ylioinas_Verbeek_Kannala_2018}
Xiaotian Li, Juha Ylioinas, Jakob Verbeek, and Juho Kannala.
\newblock Scene coordinate regression with angle-based reprojection loss for
  camera relocalization.
\newblock In {\em {ECCV Workshops}}, 2018.

\bibitem{lopez_2017}
David Lopez-Paz and Marc\textquotesingle~Aurelio Ranzato.
\newblock Gradient episodic memory for continual learning.
\newblock In {\em NeurIPS}, 2017.

\bibitem{Massiceti2017}
Daniela Massiceti, Alexander Krull, Eric Brachmann, Carsten Rother, and
  Philip~HS Torr.
\newblock Random forests versus neural networks - {What's} best for camera
  localization{?}
\newblock In {\em ICRA}, 2017.

\bibitem{MelekhovYKR17}
Iaroslav Melekhov, Juha Ylioinas, Juho Kannala, and Esa Rahtu.
\newblock Image-based localization using hourglass networks.
\newblock In {\em {ICCV Workshops}}, 2017.

\bibitem{meng2017backtracking}
Lili Meng, Jianhui Chen, Frederick Tung, James~J Little, Julien Valentin, and
  Clarence~W de Silva.
\newblock Backtracking regression forests for accurate camera relocalization.
\newblock In {\em IROS}, 2017.

\bibitem{meng2018exploiting}
Lili Meng, Frederick Tung, James~J Little, Julien Valentin, and Clarence~W de
  Silva.
\newblock Exploiting points and lines in regression forests for {RGB-D} camera
  relocalization.
\newblock In {\em IROS}, 2018.

\bibitem{nguyen_2018}
Cuong~V. Nguyen, Yingzhen Li, Thang~D. Bui, and Richard~E. Turner.
\newblock Variational continual learning.
\newblock In {\em ICLR}, 2018.

\bibitem{prabhu_2020}
Ameya Prabhu, Philip~HS Torr, and Puneet~K Dokania.
\newblock Gdumb: A simple approach that questions our progress in continual
  learning.
\newblock In {\em ECCV}, 2020.

\bibitem{radwan2018vlocnet++}
Noha Radwan, Abhinav Valada, and Wolfram Burgard.
\newblock {VLocNet++}: Deep multitask learning for semantic visual localization
  and odometry.
\newblock {\em RA-L}, 3(4):4407--4414, 2018.

\bibitem{rebuffi_2017}
Sylvestre-Alvise Rebuffi, Alexander Kolesnikov, Georg Sperl, and Christoph~H.
  Lampert.
\newblock icarl: Incremental classifier and representation learning.
\newblock In {\em CVPR}, 2017.

\bibitem{rosenbaum_2018}
Clemens Rosenbaum, Tim Klinger, and Matthew Riemer.
\newblock Routing networks: Adaptive selection of non-linear functions for
  multi-task learning.
\newblock In {\em ICLR}, 2018.

\bibitem{saputra2019distilling}
Muhamad Risqi~U Saputra, Pedro~PB de Gusmao, Yasin Almalioglu, Andrew Markham,
  and Niki Trigoni.
\newblock Distilling knowledge from a deep pose regressor network.
\newblock In {\em ICCV}, 2019.

\bibitem{sarlin2019coarse}
Paul-Edouard Sarlin, Cesar Cadena, Roland Siegwart, and Marcin Dymczyk.
\newblock From coarse to fine: Robust hierarchical localization at large scale.
\newblock In {\em CVPR}, 2019.

\bibitem{sarlin2018leveraging}
Paul-Edouard Sarlin, Fr{\'e}d{\'e}ric Debraine, Marcin Dymczyk, Roland
  Siegwart, and Cesar Cadena.
\newblock Leveraging deep visual descriptors for hierarchical efficient
  localization.
\newblock In {\em CoRL}, 2018.

\bibitem{SattlerLK11}
Torsten Sattler, Bastian Leibe, and Leif Kobbelt.
\newblock Fast image-based localization using direct {2D-to-3D} matching.
\newblock In {\em ICCV}, 2011.

\bibitem{sattler2012improving}
Torsten Sattler, Bastian Leibe, and Leif Kobbelt.
\newblock Improving image-based localization by active correspondence search.
\newblock In {\em ECCV}, 2012.

\bibitem{SattlerLK17}
Torsten Sattler, Bastian Leibe, and Leif Kobbelt.
\newblock Efficient \& effective prioritized matching for large-scale
  image-based localization.
\newblock {\em PAMI}, 39(9):1744--1756, 2016.

\bibitem{sattler2018benchmarking}
Torsten Sattler, Will Maddern, Carl Toft, Akihiko Torii, Lars Hammarstrand,
  Erik Stenborg, Daniel Safari, Masatoshi Okutomi, Marc Pollefeys, Josef Sivic,
  Fredrik Kahl, and Tomas Pajdla.
\newblock Benchmarking {6DoF} outdoor visual localization in changing
  conditions.
\newblock In {\em CVPR}, 2018.

\bibitem{Sattler17cvpr}
Torsten Sattler, Akihiko Torii, Josef Sivic, Marc Pollefeys, Hajime Taira,
  Masatoshi Okutomi, and Tomas Pajdla.
\newblock Are large-scale {3D} models really necessary for accurate visual
  localization?
\newblock In {\em CVPR}, 2017.

\bibitem{Sattler2019}
Torsten Sattler, Qunjie Zhou, Marc Pollefeys, and Laura Leal-Taixe.
\newblock Understanding the limitations of {CNN-based} absolute camera pose
  regression.
\newblock In {\em CVPR}, 2019.

\bibitem{schoenberger2016sfm}
Johannes~Lutz Sch\"{o}nberger and Jan-Michael Frahm.
\newblock Structure-from-motion revisited.
\newblock In {\em CVPR}, 2016.

\bibitem{schoenberger2016mvs}
Johannes~Lutz Sch\"{o}nberger, Enliang Zheng, Marc Pollefeys, and Jan-Michael
  Frahm.
\newblock Pixelwise view selection for unstructured multi-view stereo.
\newblock In {\em ECCV}, 2016.

\bibitem{SCoRF}
Jamie Shotton, Ben Glocker, Christopher Zach, Shahram Izadi, Antonio Criminisi,
  and Andrew Fitzgibbon.
\newblock Scene coordinate regression forests for camera relocalization in
  {RGB-D} images.
\newblock In {\em CVPR}, 2013.

\bibitem{Svarm2017}
Linus {Sv\"arm}, Olof Enqvist, Fredrik Kahl, and Magnus Oskarsson.
\newblock City-scale localization for cameras with known vertical direction.
\newblock {\em PAMI}, 39(7):1455--1461, 2016.

\bibitem{inloc}
Hajime Taira, Masatoshi Okutomi, Torsten Sattler, Mircea Cimpoi, Marc
  Pollefeys, Josef Sivic, Tom{\'{a}}s Pajdla, and Akihiko Torii.
\newblock {InLoc}: Indoor visual localization with dense matching and view
  synthesis.
\newblock In {\em CVPR}, 2018.

\bibitem{valada2018deep}
Abhinav Valada, Noha Radwan, and Wolfram Burgard.
\newblock Deep auxiliary learning for visual localization and odometry.
\newblock In {\em ICRA}, 2018.

\bibitem{valentin2016learning}
Julien Valentin, Angela Dai, Matthias Nie{\ss}ner, Pushmeet Kohli, Philip Torr,
  Shahram Izadi, and Cem Keskin.
\newblock Learning to navigate the energy landscape.
\newblock In {\em 3DV}, 2016.

\bibitem{ValentinNSFIT15}
Julien Valentin, Matthias Nie{\ss}ner, Jamie Shotton, Andrew Fitzgibbon,
  Shahram Izadi, and Philip~HS Torr.
\newblock Exploiting uncertainty in regression forests for accurate camera
  relocalization.
\newblock In {\em CVPR}, 2015.

\bibitem{vitter1985random}
Jeffrey~S Vitter.
\newblock Random sampling with a reservoir.
\newblock {\em ACM Transactions on Mathematical Software (TOMS)}, 11(1):37--57,
  1985.

\bibitem{Walch_2017_ICCV}
Florian Walch, Caner Hazirbas, Laura Leal-Taixe, Torsten Sattler, Sebastian
  Hilsenbeck, and Daniel Cremers.
\newblock Image-based localization using {LSTM}s for structured feature
  correlation.
\newblock In {\em ICCV}, 2017.

\bibitem{Yang_2019_ICCV}
Luwei Yang, Ziqian Bai, Chengzhou Tang, Honghua Li, Yasutaka Furukawa, and Ping
  Tan.
\newblock {SANet}: Scene agnostic network for camera localization.
\newblock In {\em ICCV}, 2019.

\bibitem{zhou2020kfnet}
Lei Zhou, Zixin Luo, Tianwei Shen, Jiahui Zhang, Mingmin Zhen, Yao Yao, Tian
  Fang, and Long Quan.
\newblock {KFNet}: Learning temporal camera relocalization using {Kalman}
  filtering.
\newblock In {\em CVPR}, 2020.

\bibitem{zhou2020learn}
Qunjie Zhou, Torsten Sattler, Marc Pollefeys, and Laura Leal-Taixe.
\newblock To learn or not to learn: {Visual} localization from essential
  matrices.
\newblock In {\em ICRA}, 2020.

\end{thebibliography}
}

\newpage~\newpage~

\appendix
\section*{\fontsize{18}{15}\selectfont
---Supplementary Material---}
\vspace{20px}
\section{Baseline Algorithms}
As mentioned in our main paper, the baseline algorithms \textit{Reservoir}~\cite{vitter1985random} 
and \textit{Class-balance}~\cite{chrysakis2020online} are presented in Algorithm \ref{algo:reservoir} and \ref{algo:class-balance} respectively.
\begin{algorithm}[h]
\caption{Reservoir}
\label{algo:reservoir}
\begin{algorithmic}[1] 
\STATE{ N $\leftarrow$ number of all instances observed}
\STATE{$s \sim Random(0,1) \times N$}
\IF{$s < |B|$}
\STATE{Replace a random instance in $B$ with $(x,\mathbf{y})$}
\ELSE
\STATE{Ignore}
\ENDIF
\end{algorithmic}
\end{algorithm}

\begin{algorithm}[h]
\caption{Class-balance}
\label{algo:class-balance}
\begin{algorithmic}[1] 
\IF{$c$ is not largest}
    \STATE{Select a random instance of largest class}
    \STATE{Replace it with $(x, \mathbf{y})$}
\ELSE
    \STATE{$m_c \leftarrow$ number of currently stored instances of $c$} 
    \STATE{$n_c \leftarrow$ number of total instances observed of $c$}
    \STATE{$u \sim Random(0,1)$}
    \IF{$u < m_c/n_c$}
    \STATE{Replace an instance of \textit{class}($x$) with ($x, \mathbf{y}$)}
    \ELSE
    \STATE{Ignore}
    \ENDIF
        
\ENDIF
\end{algorithmic}
\end{algorithm}

\section{Training Details}
For continual learning training purposes, we adopt two \textit{dataloaders}, namely \textit{trainloader} and \textit{bufferloader}, in our training procedure except the first scene training.  \textit{Trainloader} randomly loads data from the current scene and generates ground truth labels to feed to the model. While in \textit{bufferloader}, the training frames are loaded from our buffer and ground truth labels are either generated the same as in \textit{trainloader} or extracted from intermediate representations. The loaded image pairs $<t_i,b_i>$ are fed to the training network separately and their losses are summed together as the final loss term as shown in Eq 3 (\textit{c.f.} Sec.3.1) in the main paper, where $i$ indicates the $i^{th}$ training step.

\noindent \textbf{Rep-buff.} Since we have two classification layers and one regression layer in our training model, the loss terms in Eq 4 (\textit{c.f.} Sec.3.2) can be written as:

\begin{equation}
    L_{t} =\alpha_1 \cdot e_{l1} + \alpha_2 \cdot e_{l2}+ \beta \cdot e_{3D}
\label{eq:visual_task_loss_1}
\end{equation}

Where $e_l$ is the classification loss and $e_{3D}$ is the summed regression loss for each pixel. The weighting coefficients are set to (1,1,100000) during the training. However, in our \textit{Rep-buff}, the real-value prediction for regression layer $\tilde{y}_{3D}$ might be unbounded, and could give highly erroneous guidance to the loss term $e_{3D}$~\cite{saputra2019distilling}. Hence, instead of directly minimizing $\hat{\tilde{y}}_{3D}$ \wrt $\tilde{y}_{3D}$, as with~\cite{saputra2019distilling}, the $\tilde{y}_{3D}$ is applied as a upper bound. That is, the prediction $\hat{\tilde{y}}_{3D}$ is made as close as possible to the ground truth and not penalized if its performance surpasses $\tilde{y}_{3D}$, we have the $\beta$ in Eq \ref{eq:visual_task_loss_1} for \textit{Rep-buff}:

\begin{equation}
    \beta = \begin{cases} 100000, & \text{if $\norm[1]{\hat{\tilde{y}}_{3D} - y_{3D}}^2 > \norm[1]{\tilde{y}_{3D} - y_{3D}}^2$}; \\
    0, & \text{otherwise}.
    \end{cases}
\label{eq:visual_task_loss}
\end{equation}

\noindent \textbf{Benefits of $y_1$ over $y_{3D}$.} As mentioned in our main paper, dense 3D points $y_{3D}$ provide an accurate estimate of coverage score. However, it is computationally intensive when considering all the pixels in a candidate image. Thus, only the hierarchical cluster labels are selected to represent the dense 3D points. We have 25 labels for each level and in total 625 labels for an individual scene. In our combined scenes \textbf{i7S}, \textbf{i12S}, and \textbf{i19S}, we combine the labels tree at the first level as in ~\cite{li2020hierarchical}. That is, for \textbf{i19S}, the first level contains 475 branches. 625 $\div$ Average number of 3D points per scene is the reduced computations in such a setting.

\section{Disorder Scenes}

\begin{table}[t!]

\begin{minipage}[!t]{\columnwidth}
\hspace{5px}
\resizebox{0.9\textwidth}{!}{
\renewcommand\arraystretch{1.2}
\normalsize
\begin{tabular}{lccc}
\hline
 \multirow{2}{*}{\textbf{Scene}} & \multicolumn{3}{c}{\textbf{Accuracy (\%)}}\\
& \textbf{Reservoir} & \textbf{Class-balance} & \textbf{Buff-CS (ours)}             \\ \hline
Heads   & 78.60 & 89.20                      & {\color{red}90.90}                  \\
Kitchen    & {\color{red}47.52}                  & 46.84              & 37.18 \\
Pumpkin   & 43.65                     & 38.60                   & {\color{red}55.50} \\
Chess  & 93.40 & {\color{red}96.10}                    & 93.45                      \\
Office & {\color{red}76.14}                     & 69.65 & 69.48                      \\
Fire & 86.30 & {\color{red}93.50}                       & 92.95                     \\
Stairs  & 73.30                       & {\color{red}78.60} & 76.10                      \\ \hline
\textbf{Average}    & 71.27              & 73.21                  & {\color{red} 73.65} \\ \hline
\end{tabular}
}
\end{minipage}

\vspace{10px}

\begin{minipage}[!t]{\columnwidth}
\hspace{5px}
\resizebox{0.9\textwidth}{!}{
\renewcommand\arraystretch{1.2}
\normalsize
\begin{tabular}{lccc}
\hline
 \multirow{2}{*}{\textbf{Scene}} & \multicolumn{3}{c}{\textbf{Accuracy (\%)}}\\
& \textbf{Reservoir} & \textbf{Class-balance} & \textbf{Buff-CS (ours)}             \\ \hline
Office   & 58.85 & 56.78                     & {\color{red}60.43}                  \\
Heads    & 73.60                & 89.40            & {\color{red}89.60} \\
Fire   & 79.10                     & 81.05                  & {\color{red}84.00} \\
Chess  & 88.40 & 85.80                   & {\color{red}93.65}                     \\
Kitchen & 48.82                     & {\color{red}50.34} & 42.86                     \\
Pumpkin & 51.30 & {\color{red}59.00}                       & 54.25                    \\
Stairs  & {\color{red}80.60}              & 76.40 & 80.10                      \\ \hline
\textbf{Average}    & 68.69              & 71.25                  & {\color{red} 72.12} \\ \hline
\end{tabular}
}
\end{minipage}

\vspace{10px}

\begin{minipage}[!t]{\columnwidth}
\hspace{5px}
\resizebox{0.9\textwidth}{!}{
\renewcommand\arraystretch{1.2}
\normalsize
\begin{tabular}{lccc}
\hline
 \multirow{2}{*}{\textbf{Scene}} & \multicolumn{3}{c}{\textbf{Accuracy (\%)}}\\
& \textbf{Reservoir} & \textbf{Class-balance} & \textbf{Buff-CS (ours)}             \\ \hline
Heads   & 70.00 & 90.90                     & {\color{red}93.30}                  \\
Office    & 61.10                & 60.93              & {\color{red}63.58} \\
Chess   & 88.40                     & 86.40                   & {\color{red}93.85} \\
Kitchen  & {\color{red}47.08} & 42.56                    & 40.25                     \\
Fire & 83.45                    & 88.00 & {\color{red}87.95}                      \\
Pumpkin & 51.65 & 47.95                      & {\color{red}53.55}                     \\
Stairs  & 81.50                       & {\color{red}82.10} & 81.30                      \\ \hline
\textbf{Average}    & 69.03              & 71.26                  & {\color{red} 73.40} \\ \hline
\end{tabular}
}
\end{minipage}
\vspace{10px}
\caption{The accuracy on individual scenes of \textbf{i7S} with different training order  after  the  training  is  complete.  The order for the last scene is preserved. All methods employs an \textit{Img-buff} buffer of size 256.The best results are highlighted in red.} 
\label{tab:disorder}

\end{table}

At this point, we provide the results for \textbf{i7S}  where models are trained with three different permutations of scenes, with buffer size $B = 256$ and \textit{Img-buff} replay. Note that we preserve the order of the last scene. 
Table~\ref{tab:disorder} present the accuracy on the test images of individual scenes and overall accuracy on \textbf{i7S}. The results show that our method is not affected by the order of scenes and performs comparably to or exceeds the baselines as mentioned in the main paper.

\section{Rep-buff}
\begin{table}[t!]

\resizebox{0.45\textwidth}{!}{
\renewcommand\arraystretch{1.2}
\normalsize

\begin{tabular}{lccc}
\hline
\multirow{2}{*}{\textbf{Scenes}} & \multicolumn{3}{c}{\textbf{Accuracy (\%)}}                                                                 \\
                                 & \textbf{Weights = 0,0,1} & \textbf{Weights = 0,0,1e5} & \textbf{Weights = 1,1,0} \\ \hline
apt1/kitchen     & 84.59 & 77.87 & 82.91 \\
apt1/living      & 77.28 & 74.65 & 79.11 \\
apt2/bed         & 85.29 & 78.92 & 80.88 \\
apt2/kitchen     & 96.19 & 95.71 & 97.62 \\
apt2/living      & 79.08 & 80.23 & 77.65 \\
apt2/luke        & 46.79 & 42.15 & 45.83 \\
office1/gates362 & 67.88 & 64.25 & 72.02 \\
office1/gates381 & 50.43 & 51.57 & 56.41 \\
office1/lounge   & 87.46 & 86.54 & 86.54 \\
office1/manolis  & 81.18 & 79.42 & 74.65 \\
office2/5a       & 94.16 & 94.77 & 93.96 \\
office2/5b       & 97.28 & 96.26 & 97.04 \\ \hline
\textbf{Average}          & 78.97 & 76.86 & 78.72 \\ \hline
\end{tabular}
}

\vspace{5px}
\caption{The accuracy on individual scenes of \textbf{i12S} with buffer size $B = 128$ and different assigned weights on \textit{Rep-buff}.} 
\label{tab:weights}
\end{table}
In all our experiments the number of cluster levels used is $L = 2$. The corresponding network predictions at each cluster level, $\hat{y_1}, \hat{y_2}$ and the final 3D coordinates $\hat{y_{3D}}$ are stored in \textit{Rep-buff}. We analyze the influence of experience-replay using intermediate predictions $\hat{y_1}, \hat{y_2}$ and the final layer predictions $\hat{y_{3D}}$ in this section.

The different layers are assigned weights, $\alpha_{1},  \alpha_{2}, \beta $. For ($\alpha_{1},  \alpha_{2}, \beta $) = (1,1,0) and (0,0,1) respectively, results in table~\ref{tab:weights} show that both these conditions obtain comparable performance. For \textit{Buff-CS}, the results are better than the weight distribution (1,1,100000) originally proposed by~\cite{li2020hierarchical}.

\section{Additional Results on i12S}

\begin{figure*}
\begin{center}

   \includegraphics[width=1\linewidth]{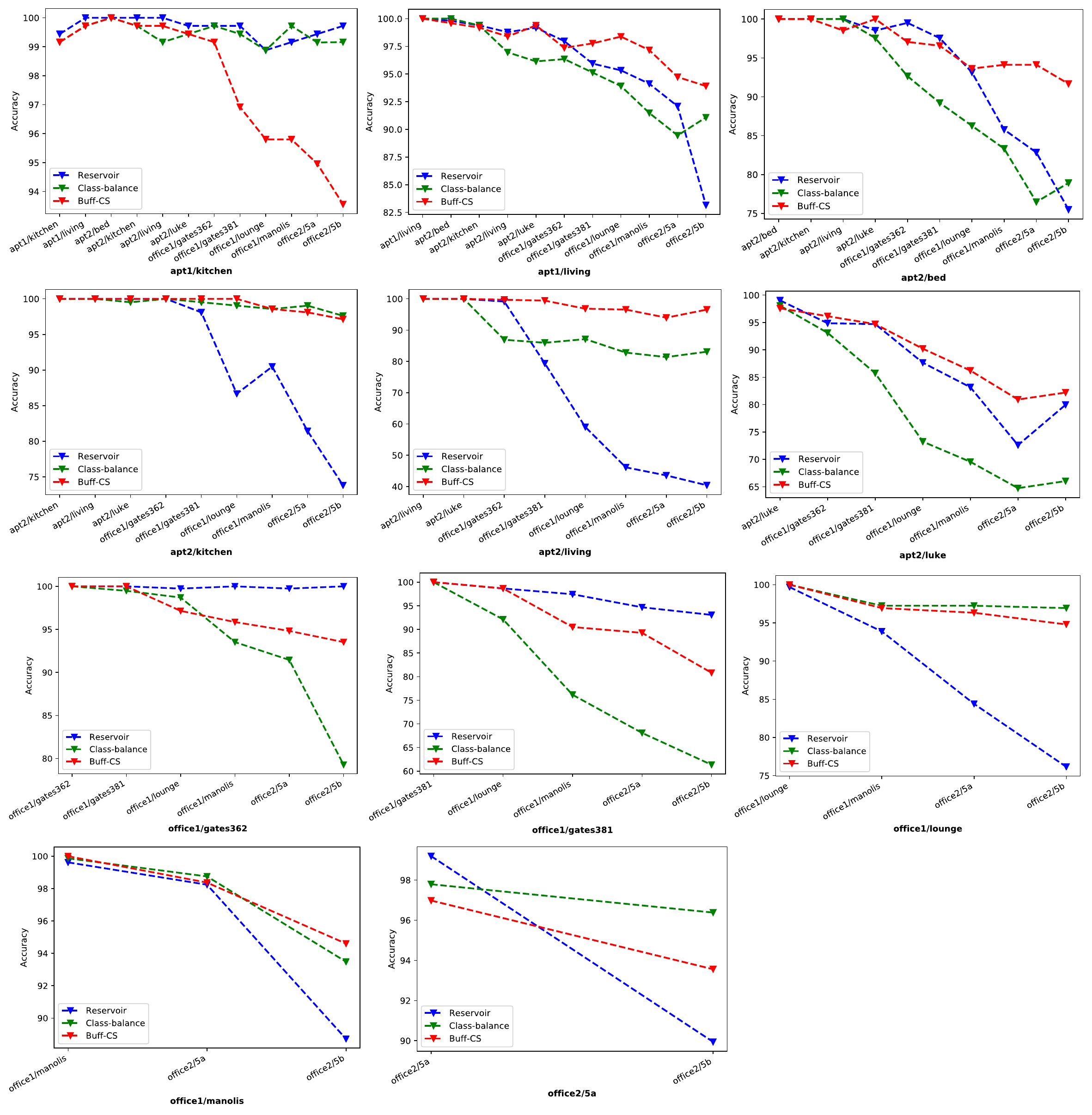}
   \end{center}
   \caption{The accuracy (error $< 5$ cm, $5^\circ$) on individual scenes of \textbf{i12S} (except for the last scene) at each stage of the training. The x axis indicates the training progress. All methods employs an \textit{Img-buff} buffer of size 256. The results show for majority of scenes \textit{Buff-CS} outperforms \textit{Class-balance} and \textit{Reservoir}.}
        \label{fig:i12_accuracy}
\vspace{-5px}
\end{figure*}

Similar to the \textbf{i7S} results presented in the main paper, we show the results on individual scenes  of \textbf{i12S} after training is completed in Table~\ref{tab:i12_results}, and the average accuracy over different stages of the training process on each scene of \textbf{i12S} in Table~\ref{tab:i12_average}. In addition, the accuracy on individual scenes of \textbf{i12S} at each stage of the training is provided in Fig.~\ref{fig:i12_accuracy}. The results show for majority of scenes \textit{Buff-CS} outperforms \textit{Class-balance} and \textit{Reservoir} across different evaluation settings.

\begin{table}[b]

        \resizebox{0.45\textwidth}{!}{
        \renewcommand\arraystretch{1.2}
        \normalsize
    
            \begin{tabular}{lccc}
            \hline
                     & \multicolumn{3}{c}{\textbf{Accuracy (\%)}}                                                      \\
            \multirow{-2}{*}{\textbf{Scenes}} & \textbf{Reservoir} & \textbf{Class-balance} & \textbf{Buff-CS(ours)} \\ \hline
             apt1/kitchen     & 97.68                        & {\color{red} 98.60}  & 93.84                        \\
             apt1/living      & 81.95                        & 89.05                        & {\color{red} 95.74} \\
             apt2/bed         & 74.02                        & 73.53                        & {\color{red} 87.25} \\
             apt2/kitchen     & 71.90                         & 98.09                        & {\color{red} 98.57} \\
             apt2/living      & 41.54                        & 81.38                        & {\color{red} 94.27} \\
             apt2/luke        & 80.93                        & 60.10                         & {\color{red} 84.54} \\
             office1/gates362 & {\color{red} 100}   & 82.38                        & 91.45                        \\
             office1/gates381 & {\color{red} 93.44} & 62.77                        & 76.26                        \\
             office1/lounge   & 71.86                        & {\color{red} 96.02} & 93.88                        \\
             office1/manolis  & 87.58                        & 92.72                        & {\color{red} 94.60}  \\
             office2/5a       & 92.35                        & {\color{red} 98.19} & 94.97                        \\
             office2/5b       & {\color{red} 97.28} & 96.05                        & 96.79                        \\ \hline
             \textbf{Average}          & 82.54                        & 85.72                        & {\color{red} 91.85} \\ \hline
             \end{tabular}
         }
         \vspace{5px}
         {
         \caption{The percentage of accurately localized test images (error $<$ 5 cm, $5^\circ$) on \textbf{i12S} with the buffer size $B = 256$,  after  the  training  is  complete. Here we use \textit{Img-buff} for replay. The best results are highlighted in red. }
         \label{tab:i12_results}}
\end{table}

\begin{table}
        \resizebox{0.45\textwidth}{!}{
        \renewcommand\arraystretch{1.2}
        \normalsize
        
        \begin{tabular}{lccc}
        \hline
                         & \multicolumn{3}{c}{\textbf{Average Accuracy (\%)}}                                                      \\
        \multirow{-2}{*}{\textbf{Scenes}} & \textbf{Reservoir} & \textbf{Class-balance} & \textbf{Buff-CS(ours)} \\ \hline
        apt1/kitchen     & {\color{red} 99.65}                       & 99.44  & 97.83                        \\
        apt1/living      & 95.98                       & 95.44                        & {\color{red} 97.81} \\
        apt2/bed         & 93.28                        & 90.44                        & {\color{red} 96.57} \\
        apt2/kitchen     & 92.28                         & 99.26                       & {\color{red} 99.31} \\
        apt2/living      & 70.95                        & 88.41                       & {\color{red} 97.89} \\
        apt2/luke        & 87.43                       & 78.64                        & {\color{red} 89.72} \\
        office1/gates362 & {\color{red} 99.91}   & 93.74                        & 96.89                       \\
        office1/gates381 & {\color{red} 96.77} & 79.54                       & 91.85                       \\
        office1/lounge   & 88.53                        & {\color{red} 97.86} & 97.02                        \\
        office1/manolis  & 95.52                        & 97.37                        & {\color{red} 97.66}  \\
        office2/5a       & 94.57                        & {\color{red} 97.09} & 95.27                        \\
        office2/5b       & {\color{red} 99.51} & 97.04                        & 94.07                       \\ \hline
        \textbf{Total Average}          & 92.83                        & 92.86                        & {\color{red} 95.99} \\ \hline
        \end{tabular}
        }
        \vspace{5px}
        {
        \caption{The average accuracy over different stages of the training process on each scene of \textbf{i12S} with the buffer size $B = 256$ and \textit{Img-buff} replay. Our method has overall better performance compared to the other two methods.} 
        \label{tab:i12_average}
        }

\end{table}

\end{document}